\documentclass[10pt,twocolumn,letterpaper]{article}
\pdfoutput=1
\usepackage{iccv}              
\usepackage{algorithm,algorithmic}
\usepackage{enumitem}
\definecolor{iccvblue}{rgb}{0.21,0.49,0.74}
\usepackage[pagebackref,breaklinks,colorlinks,allcolors=iccvblue]{hyperref}

\def\paperID{13168} 
\def\confName{ICCV}
\def\confYear{2025}

\title{HQC-NBV: A Hybrid Quantum-Classical View Planning Approach}

\author{Xiaotong Yu\\
The Hong Kong Polytechnic University\\
Hung Hom, Hong Kong SAR\\
{\tt\small xiaotong.yu@connect.polyu.hk}
\and
Chang Wen Chen\\
The Hong Kong Polytechnic University\\
Hung Hom, Hong Kong SAR\\
{\tt\small changwen.chen@polyu.edu.hk}
}

\begin{document}
\maketitle
\begin{abstract}
Efficient view planning is a fundamental challenge in computer vision and robotic perception, critical for tasks ranging from search and rescue operations to autonomous navigation. While classical approaches, including sampling-based and deterministic methods, have shown promise in planning camera viewpoints for scene exploration, they often struggle with computational scalability and solution optimality in complex settings. This study introduces HQC-NBV, a hybrid quantum-classical framework for view planning that leverages quantum properties to efficiently explore the parameter space while maintaining robustness and scalability. We propose a specific Hamiltonian formulation with multi-component cost terms and a parameter-centric variational ansatz with bidirectional alternating entanglement patterns that capture the hierarchical dependencies between viewpoint parameters. Comprehensive experiments demonstrate that quantum-specific components provide measurable performance advantages. Compared to the classical methods, our approach achieves up to 49.2\% higher exploration efficiency across diverse environments. Our analysis of entanglement architecture and coherence-preserving terms provides insights into the mechanisms of quantum advantage in robotic exploration tasks. This work represents a significant advancement in integrating quantum computing into robotic perception systems, offering a paradigm-shifting solution for various robot vision tasks.
\end{abstract}
    
\section{Introduction}
\label{sec:intro}
In unknown scene perception, determining where to move a camera next - known as the informative view planning problem - can mean the difference between success and failure in critical applications. For instance, in search and rescue operations, inefficient view planning can lead to crucial delays, where every minute matters for survival rates. Similar challenges exist in autonomous navigation and robotic manipulation, where systematic and efficient environment exploration directly impacts task completion time and resource utilization. 
The Next Best View (NBV) problem represents a fundamental challenge in computer vision and robotic exploration and perception, where the objective is to determine optimal sequential viewpoints to maximize information gained about the environment with each move. Solving the NBV problem effectively can significantly enhance the performance of robotic systems by ensuring that they gather the most relevant and useful visual data with minimal resources. 
\begin{figure}[t]
    \centering
    \includegraphics[width=0.75\linewidth]{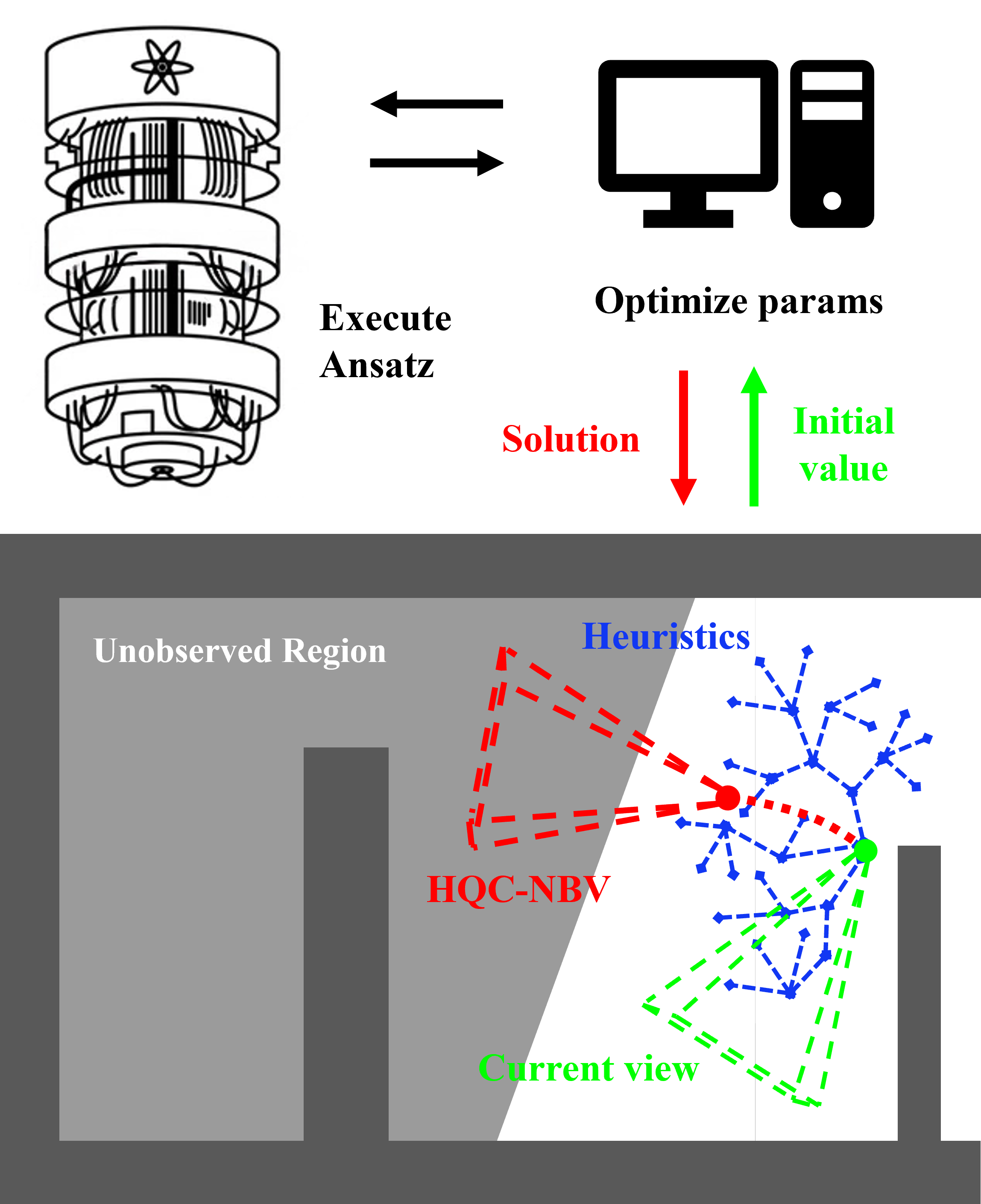}
    \caption{Execution logic of our HQC-NBV. Different from the classical approaches, we do not rely on heuristics but leverage quantum superposition to simultaneously evaluate multiple view parameters and quantum entanglement to capture complex dependencies between movement decisions.}
    \label{fig:cover}
\end{figure}

Next-Best-View was initially introduced for exploring unknown areas using mobile robots \cite{bircher2016receding, meng2017two, selin2019efficient, xu2021autonomous}. Early approaches can be primarily categorized into sampling-based and deterministic methods. Sampling-based approaches \cite{bircher2016receding, respall2021fast, batinovic2022shadowcasting} typically employ Rapidly-exploring Random Trees (RRT) or RRT* within known free space \cite{lavalle1998rapidly, karaman2011sampling}, generating candidate views and selecting the optimal one based on information gain versus cost metrics. While these methods have shown success in simple environments, they face significant scalability challenges in complex scenarios, often requiring exponentially increasing computational resources with environment size. Deterministic methods \cite{xiang2022combined, zaenker2023graph, steinbrink2021rapidly}, on the other hand, rely on heuristics to guide viewpoint selection, focusing on specific metrics or uncertainty minimization. Despite their convenient and widespread deployment in real-world mobile platforms, these classical approaches suffer from fundamental limitations. Heuristic-based methods often struggle to find global optima, particularly in large-scale environments, while sampling techniques frequently result in suboptimal solutions due to their approximative nature of the solution space. 

To address these challenges, we explore the potential of quantum computing in solving the NBV problem. Quantum computing has recently demonstrated promising results in various computer vision tasks, including multi-model fitting \cite{farina2023quantum}, multi-object tracking \cite{zaech2022adiabatic}, motion segmentation \cite{arrigoni2022quantum}, and graph matching \cite{benkner2020adiabatic, benkner2021qm}. The quantum advantage derives from its ability to leverage quantum phenomena such as superposition and entanglement, enabling efficient exploration of vast solution spaces. The NBV problem is particularly well-suited for quantum approaches due to its combinatorial nature and the presence of complex parameter interdependencies that can be naturally encoded in quantum entanglement structures. Recent quantum implementations in the related problem \cite{farina2023quantum} have shown the potential of adiabatic quantum computing (AQC) in disjoint set cover problems, suggesting a similar potential for view planning optimization. 

Our work introduces a novel hybrid quantum-classical framework that combines the computational advantages of quantum systems with the robustness of classical optimization techniques, as shown in Figure \ref{fig:cover}. This hybrid approach aims to overcome the limitations of traditional methods while maintaining collocation with current quantum computing devices. Specifically, our contributions are as follows: 
\begin{itemize}
    \item A novel hybrid quantum-classical approach for informative view planning featuring a Hamiltonian formulation of the NBV problem that effectively maps robotic navigation intuition into the quantum computing paradigm.
    \item A parameter-centric variational ansatz design with bidirectional alternating entanglement patterns that capture the hierarchical dependencies between view parameters, allowing simultaneous exploration of movement directions, distances, and orientations.
    \item Comprehensive experimental validations demonstrate the contribution of quantum-specific components (i.e., entanglement architecture and coherence-preserving terms), as well as the robustness and effectiveness of our approach in different experimental settings, achieving 7.9-49.2\% higher exploration efficiency compared to the classical methods. 
\end{itemize}
To the best of our knowledge, this is the first study to propose a hybrid quantum-classical approach for informative view planning, opening new possibilities for efficient robot perception and navigation.
The paper content is organised as follows: Section \ref{sec:related} and Section \ref{prelimi} provide an overview of the related work and essential quantum computing preliminaries. Section \ref{meth} presents the problem formulation of NBV and our proposed hybrid quantum-classical approach. Section \ref{impl} presents the implementation of the proposed framework and an additional local strategy. The experimental setup and results are presented in Section \ref{expAres}. Finally, we analyse the results and draw the conclusion in Section \ref{conc}.

\section{Related Work}
\label{sec:related}
\subsection{Informative View Planning}

Informative view planning, particularly in non-model-based visual acquisition scenarios where no prior environmental knowledge is available, requires real-time decision-making for each viewpoint. This planning process has evolved significantly since its introduction in the 1980s \cite{connolly1985determination, maver1993occlusions}, with approaches generally falling into three main categories: surface-based, volumetric, and hybrid methods.
Surface-based approaches represent the 3D environment as a mesh and evaluate viewpoints by analyzing the mesh surface \cite{chen2005vision}. For instance, Krainin et al. \cite{karaman2011sampling} modelled uncertainty using Gaussian distributions along camera rays and quantified information gain through entropy reduction weighted by surface area. While these methods enable direct quality assessment during reconstruction and can handle dynamic objects accurately \cite{sui2022accurate}, they are computationally intensive due to complex visibility calculations \cite{scott2003view}.
Volumetric methods, alternatively, employ voxel-based representations that simplify visibility calculations and occupation probability estimation \cite{hornung2013octomap}. These methods evaluate potential views by ray-casting through voxel space, offering computational efficiency at the cost of direct surface modelling capability. To leverage the advantages of both approaches, hybrid methods \cite{kriegel2015efficient} combine surface and volumetric representations, achieving a balance between accuracy and efficiency.
The Next-Best-View (NBV) paradigm has emerged as a dominant strategy in informative view planning, iteratively selecting viewpoints to maximize information gain. Information gain metrics have evolved from simple unknown voxel counting \cite{banta2000next} to sophisticated measures such as information theoretic entropy \cite{kriegel2015efficient} and proximity-based volumetric information \cite{delmerico2018comparison}. A significant advancement came with Bircher et al. \cite{bircher2016receding} receding horizon NBV (RH-NBV) approach, which incorporated model predictive control principles to avoid local minima through selective path execution and distance-based penalization. Recent developments include ratio-based utility functions \cite{schmid2020efficient} and uncertainty-guided schemes \cite{jin2023neu} that enhance reconstruction accuracy.
While frontier-based methods \cite{yamauchi1997frontier} offer an alternative approach by focusing on boundaries between explored and unexplored areas, particularly effective in high-speed flight scenarios \cite{cieslewski2017rapid, batinovic2021multi}, they lack the flexible information gain formulation characteristic of NBV methods. Recent work has extended view planning to specialized applications, such as fruit mapping \cite{menon2023nbv}, demonstrating the adaptability of these approaches to diverse scenarios.

\subsection{Quantum Computer Vision}

There is a growing interest in the potential of quantum computing for solving challenging problems in computer vision. The inherent advantages of quantum systems, e.g. superposition, entanglement, and quantum parallelism, offer unique opportunities to tackle computational intensive tasks more effectively.
Farina et al. \cite{farina2023quantum} propose Quantum Unconstrained Multi-Model Fitting (QUMF and DEQUMF) method effectively utilizes quantum annealing to optimize the selection of multiple geometric models as a combinatorial optimization, taking advantage of quantum superposition to explore multiple solutions simultaneously. Zaech et al. \cite{zaech2022adiabatic} map the multi-object tracking problem to an Ising model and utilize adiabatic quantum computing (AQC) to find optimal assignments through a quadratic unconstrained binary optimization (QUBO) approach. Similarly, Arrigoni et al. \cite{arrigoni2022quantum} reformulate the motion segmentation problem into a quadratic unconstrained binary optimization format suitable for adiabatic quantum computing. Benkner et al. \cite{benkner2020adiabatic} reformulate quadratic assignment problems (QAPs) with permutation matrix constraints into a quadratic unconstrained binary optimization format suitable for quantum annealing. Later, they present an iterative method for solving the quadratic assignment problem in shape matching using quantum annealing, achieving high-quality correspondences between non-rigidly transformed shapes \cite{benkner2021qm}. The existing studies focus on reformulating the problem to the Ising model or quadratic unconstrained binary optimization problems and solving them by adiabatic quantum computing. With the recent development of quantum technologies, we are currently within the Noisy Intermediate-Scale Quantum (NISQ) era \cite{lau2022nisq}, where new possibilities are emerging for solving complex problems using variational quantum algorithms, quantum approximate optimization algorithms, etc.

\section{Quantum Computing Preliminaries}
\label{prelimi}

\subsection{Basic Concepts and Properties}

\textbf{Quantum bit (qubit)} is the basic computational element in quantum computers. Different from the classical bit, a qubit has the state of a superposition formed by two basis states $|0\rangle=[1\ 0]^T$ and $|1\rangle=[0\ 1]^T$. Qubits can be prepared via different kinds of approaches, including but not limited to photons, trapped ions, Si-based quantum dots, and superconducting circuits. In the NISQ era, the most widely used one is the superconducting circuit approach, leveraging its advantage in scalability. 

\textbf{Superposition} refers to the property of the quantum state that can be a linear combination of the corresponding basis states. i.e. a qubit state $|\psi\rangle$ can be described as:
\begin{equation}
    |\psi\rangle=c_1|0\rangle+c_2|1\rangle
\end{equation}
$c_1$ and $c_2$ are complex numbers, named probability amplitudes, with $|c_1|^2+|c_2|^2=1$.

\textbf{Entanglement} is the critical property for quantum computing. In an entangled system, the state of each qubit is interconnected with the states of the other qubits without space limit, meaning that no qubit can be described independently of the rest of the system. 

\textbf{Measurement} the state of a qubit yields one of the basis states, either $|0\rangle$ or $|1\rangle$. The probability of measuring $|0\rangle$ and $|1\rangle$ are given by $|c_1|^2$ and $|c_2|^2$ respectively. Once a measurement is performed, it corresponds to an observation of the qubit, leading to the collapse of its wave function.

\subsection{NISQ and AQC}
Adiabatic Quantum Computing (AQC) is a paradigm that focuses on solving optimization problems by evolving a quantum system from a known initial state to a final state encoding the solution. This is achieved by slowly evolution the system's Hamiltonian $H(t)$, ensuring the ground state of the initial Hamiltonian $H_0$ evolves to the ground state of the final Hamiltonian $H_f$.
\begin{equation}
    H(t)=(1-t/T)H_0+(t/T)H_f
\end{equation}
The Noisy Intermediate-Scale Quantum (NISQ) era represents the current stage of quantum technology, characterized by quantum computers with a moderate number of qubits (typically a few dozen to a few hundred) that are prone to noise and errors. Despite these limitations, NISQ devices show promise in solving practical problems using variational quantum algorithms and quantum approximate optimization algorithms, which are resilient to noise and can be implemented on current hardware. Variational quantum algorithms (VQAs) are a class of hybrid quantum-classical algorithms designed to work on NISQ devices. They use a parameterized quantum circuit $U(\theta)$ to prepare a quantum state $|\psi(\theta)\rangle$ and then measure an observable $O$. The goal is to minimize the expectation value $E(\theta)$ of a given Hamiltonian $H$:
\begin{equation}
    E(\theta)=\langle\psi(\theta)|H|\psi(\theta)\rangle
\end{equation}
The parameters $\theta$ are optimized using a classical optimizer to find the minimum energy state.


\section{Methodology}
\label{meth}
\begin{figure*}
    \centering
    \includegraphics[width=1.0\linewidth]{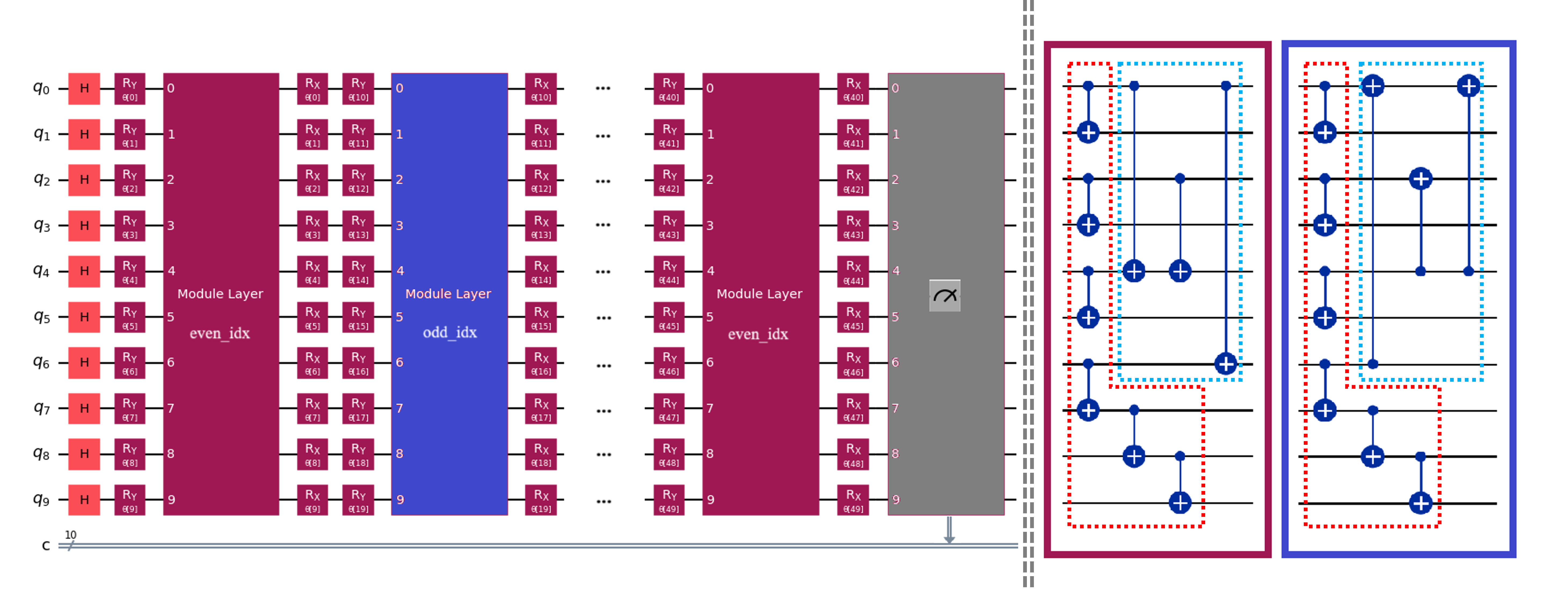}
    \caption{(Left) The block scheme of the proposed variational ansatz; (Right) The block module with even and odd index. The light red dash line highlights the intra-group entanglement, and the light blue dash line highlights the bidirectional inter-group entanglement.}
    \label{fig:meth_overview}
\end{figure*}
\subsection{Problem Formulation}
Consider a bounded 2D environment $S\subset \mathbb{R}^2$ containing a set of obstacles $\mathcal{O}=\{O_1, O_2, ..., O_M\}$ to be explored. The camera's viewpoint is given by $v=(x,y,\theta)\in \mathcal{C}$, where $(x,y)$ denotes the position and $\theta$ denotes the orientation. We aim to find the next best viewpoint that maximizes the perception of the unexplored region while minimizing the movement cost and ensuring physical continuity. It can be formulated as an optimization problem: 
\begin{equation}
    \min_{v \in \mathcal{C}} J(v) = -E(v) + \lambda_m M(v) + \lambda_s S_a(v)
\end{equation}
subject to:
\begin{equation}
    P(v' \rightarrow v) \cap \mathcal{O} = \emptyset
\end{equation}
where $E(v)$ denotes the exploration benefit function quantifying potential information gain, $M(v)$, $S_a(v)$ denote the movement cost and the safety cost. $\lambda_m$ and $\lambda_s$ are the weight parameters. And $P(v' \rightarrow v)$ represents the path between viewpoints. The exploration benefit function $E(v)$ measures the amount of new information gained by moving to viewpoint $v$, while $M(v)$ and $S_a(v)$ penalize excessive movement and proximity to obstacles. To evaluate $E(v)$ based on the historical and current views, we maintain an observed grid map $\mathcal{M}$ representing the accumulated knowledge of the environment, with each grid in a ternary state: unknown, free space, or occupied.

\subsection{Proposed Method}
\subsubsection{Problem Hamiltonian Formulation}

We formulate the informative view planning problem as a combinatorial optimization task through a carefully designed Hamiltonian $\hat{H}$. This Hamiltonian is constructed as a weighted sum of Pauli operators, where each term encodes specific aspects of the exploration problem:
\begin{equation}
\hat{H} = \sum_{i} \alpha_i \hat{P}_i
\end{equation}
Here, $\hat{P}_i$ represents a Pauli string (tensor product of Pauli matrices $I$, $X$, $Y$, $Z$), and $\alpha_i$ is the corresponding coefficient that determines the strength and direction of each term's contribution to the optimization objective. We decompose the Hamiltonian into five functional components:
\begin{equation}
\hat{H} = \hat{H}_{\text{dir}} + \hat{H}_{\text{dist}} + \hat{H}_{\text{adj}} + \hat{H}_{\text{orient}} + \hat{H}_{\text{coh}}
\end{equation}

Our novel approach uses a 10-qubit system to encode viewpoint parameters for efficient exploration:
\begin{itemize}
    \item 2 qubits for the main movement direction
    \item 2 qubits for relative movement distance
    \item 2 qubits for directional tuning
    \item 4 qubits for the view orientation angle
\end{itemize}
This allocation of qubits is carefully designed to match the physical parameters' importance and range requirements. The main direction is encoded with the first 2 qubits. Distance and adjustment parameters are allocated 2 qubits each, providing sufficient precision for movement magnitudes while keeping the quantum circuit complexity manageable. The camera orientation angle receives 4 qubits to precisely direct the field of view toward information-rich regions. This allocation reflects the hierarchical nature of the exploration task while maintaining an efficient quantum representation with only 10 qubits total.

The directional component $\hat{H}_{\text{dir}}$ encodes exploration preferences using $Z$ operators on the first two qubits:
\begin{equation}
\hat{H}_{\text{dir}} = \alpha_{Z_1} (Z \otimes I \otimes \cdots) + \alpha_{Z_2} (I \otimes Z \otimes \cdots) + \alpha_{ZZ} (Z \otimes Z \otimes \cdots)
\end{equation}
where the coefficients are derived from directional exploration values (representing the unexplored density in each cardinal direction, where $E$, $N$, $W$, and $S$ correspond to East, North, West, and South directions, respectively):
\begin{equation}
\begin{aligned}
    \alpha_{Z_1} &= \lambda_{\text{dir}} \cdot \tanh(e_W + e_S - e_E - e_N) \\ 
    \alpha_{Z_2} &= \lambda_{\text{dir}} \cdot \tanh(e_N + e_S - e_E - e_W) \\
    \alpha_{ZZ} &= \lambda_{\text{diag}} \cdot \tanh(e_{SE} + e_{NE} - e_{SW} - e_{NW})
\end{aligned}
\end{equation}
 $\alpha_{ZZ}$ captures directional interdependencies. Here $\lambda_{\text{dir}}$ and $\lambda_{\text{diag}}$ are weighting parameters.

The distance component $\hat{H}_{\text{dist}}$ controls movement magnitude using $Z$ operators on distance-encoding qubits:
\begin{equation}
\hat{H}_{\text{dist}} = \sum_{i=0}^{q_p-3} \alpha_{Z_{\text{dist},i}} (I^{\otimes l_{\text{dist}}+i} \otimes Z \otimes I^{\otimes n-l_{\text{dist}}-i-1})
\end{equation}
with coefficients proportional to observed obstacle proximity and bit significance:
\begin{equation}
\alpha_{Z_{\text{dist},i}} = \lambda_{\text{dist}} \cdot 2^{-(i+1)} \cdot \frac{d_{\text{obs}}}{d_{\text{max}}}
\end{equation}

The adjustment component $\hat{H}_{\text{adj}}$ provides fine-tuning of the movement direction through $Z$ operators on adjustment-encoding qubits:
\begin{equation}
\hat{H}_{\text{adj}} = \sum_{i=0}^{q_p-3} \alpha_{Z_{\text{adj},i}} (I^{\otimes l_{\text{adj}}+i} \otimes Z \otimes I^{\otimes n-l_{\text{adj}}-i-1})
\end{equation}
where the coefficients are scaled by both bit significance and the remaining unexplored area:
\begin{equation}
\alpha_{Z_{\text{adj},i}} = \lambda_{\text{adj}} \cdot 2^{-(i+1)} \cdot (1-c)
\end{equation}
Here, $(1-c)$ represents the proportion of unexplored environment, ensuring that directional adjustments become more precise as exploration progresses. The exponential term $2^{-(i+1)}$ maintains the binary significance hierarchy, with higher-order bits contributing more substantially to the directional refinement.

The orientation component $\hat{H}_{\text{orient}}$ combines target direction terms with exploration-promoting terms:
\begin{equation}
\hat{H}_{\text{orient}} = \sum_{i=0}^{q_p-1} \alpha_{Z_{\text{orient},i}} {Z}_i + \sum_{i=0}^{q_p-1} \alpha_{X_{\text{orient},i}} {X}_i + \sum_{i,j} \alpha_{ZZ_{\text{couple}}} {Z}_i{Z}_j
\end{equation}
where ${Z}_i$ and ${X}_i$ represent $Z$ and $X$ operators acting on orientation qubits. The target direction coefficients are:
\begin{equation}
\alpha_{Z_{\text{orient},i}} = \lambda_{\text{orient}} \cdot 2^{-(i+1)} \cdot \rho \cdot (1-D) \cdot b_i
\end{equation}
where $\rho$ represents normalized point density, $D$ is angular dispersion, and $b_i \in \{-1,1\}$ encodes the target angle. For high dispersion scenarios, the exploration coefficients are:
\begin{equation}
\alpha_{X_{\text{orient},i}} = \lambda_{\text{orient-X}} \cdot D \cdot (\gamma)^i
\end{equation}
where $\lambda_{\text{orient-X}}$ is a weighting parameter and $\gamma$ is a decay factor for higher-order bits.
Finally, the coherence component $\hat{H}_{\text{coh}}$ maintains quantum advantage through $X$ operators and entangling terms:
\begin{equation}
\hat{H}_{\text{coh}} = \sum_{i} \alpha_{X_i} {X}_i + \sum_{(i,j) \in \mathcal{P}} \alpha_{XX_{i,j}} {X}_i{X}_j
\end{equation}
where $\mathcal{P}$ represents selected qubit pairs (based on the physical meaning and correlation between view parameters. i.e., direction-adjustment, distance-adjustment, direction-orientation). The entanglement coefficients scale with unexplored area:
\begin{equation}
\alpha_{XX_{i,j}} = \lambda_{\text{coh}} \cdot (1-c)
\end{equation}
This Hamiltonian structure enables the quantum system to simultaneously evaluate multiple movement strategies while encoding complex spatial relationships and exploration priorities. By minimizing the expectation value $\langle \psi|\hat{H}|\psi \rangle$, we identify quantum states that correspond to optimal next viewpoints for efficient environment exploration.

\subsubsection{Variational Ansatz Design}

We develop a multi-layered parameterized quantum circuit $U(\vec{\theta})$ that acts on $n$ qubits initialized in a uniform superposition state:
\begin{equation}
|\psi(\vec{\theta})\rangle = U(\vec{\theta})|+\rangle^{\otimes n}
\end{equation}
where $|+\rangle^{\otimes n}$ represents the uniform superposition state obtained by applying Hadamard gates to all qubits in the $|0\rangle^{\otimes n}$ state. The total number of qubits is determined by the parameter encoding scheme:
\begin{equation}
n = 2 + 2(q_p - 2) + q_p
\end{equation}
where $q_p$ is the number of qubits allocated per parameter, with $q_p = 4$ in our implementation.
The circuit architecture consists of $L = 5$ alternating layers of parameterized rotations and structured entanglement operations:
\begin{equation}
U(\vec{\theta}) = U_L(\vec{\theta}_L) \cdots U_2(\vec{\theta}_2) U_1(\vec{\theta}_1)
\end{equation}
Each layer $U_l(\vec{\theta}_l)$ comprises three key components:
\begin{equation}
U_l(\vec{\theta}_l) = U_l^{\text{rot}}(\vec{\theta}_l^{\text{rot}}) \cdot U_l^{\text{ent}} \cdot U_l^{\text{rx}}(\vec{\theta}_l^{\text{rx}})
\end{equation}

The rotational component $U_l^{\text{rot}}(\vec{\theta}_l^{\text{rot}})$ applies $R_y$ rotations to encode the parameters into the quantum state. These rotations are partitioned according to the parameter groups:
\begin{equation}
\begin{split}
    U_l^{\text{rot}}(\vec{\theta}_l^{\text{rot}}) = & \bigotimes_{i=0}^{1} R_y(\theta_{l,i}^{\text{dir}}) \otimes \bigotimes_{i=0}^{q_p-3} R_y(\theta_{l,i}^{\text{dist}}) \otimes \bigotimes_{i=0}^{q_p-3} R_y(\theta_{l,i}^{\text{adj}}) \\
    & \otimes \bigotimes_{i=0}^{q_p-1} R_y(\theta_{l,i}^{\text{orient}})
\end{split}
\end{equation}
This structured encoding allows the circuit to independently modulate each parameter while maintaining correlations through subsequent entanglement operations.

The entanglement component $U_l^{\text{ent}}$ establishes quantum correlations between qubits following a two-level hierarchical strategy, as is shown in Figure \ref{fig:meth_overview}: intra-group entanglement followed by inter-group entanglement. The intra-group entanglement creates linear chains of CNOT gates within each parameter group:
\begin{equation}
U_l^{\text{intra}} = \prod_{g \in \{\text{dir}, \text{dist}, \text{adj}, \text{orient}\}} \prod_{i=l_g}^{l_g+n_g-2} \text{CNOT}_{i,i+1}
\end{equation}
where $l_g$ and $n_g$ represent the starting position and size of group $g$, respectively.

The inter-group entanglement establishes connections between parameter groups, with the pattern alternating between even and odd layers:
\begin{equation}
U_l^{\text{inter}} = 
\begin{cases}
\text{CNOT}_{l_{\text{dir}},l_{\text{adj}}} \cdot \text{CNOT}_{l_{\text{dist}},l_{\text{adj}}} \cdot \text{CNOT}_{l_{\text{dir}},l_{\text{orient}}},\ l\%2=0 \\
\text{CNOT}_{l_{\text{orient}},l_{\text{dir}}} \cdot \text{CNOT}_{l_{\text{adj}},l_{\text{dist}}} \cdot \text{CNOT}_{l_{\text{adj}},l_{\text{dir}}},\ l\%2\neq0
\end{cases}
\end{equation}
This bidirectional entanglement pattern ensures information flow between parameter groups in both forward and reverse directions, facilitating complex correlations while maintaining circuit depth efficiency.

The final component in each layer applies $R_x$ rotations to all qubits:
\begin{equation}
U_l^{\text{rx}}(\vec{\theta}_l^{\text{rx}}) = \bigotimes_{i=0}^{n-1} R_x(\theta_{l,i}^{\text{rx}})
\end{equation}
These rotations introduce non-commutativity with respect to the $R_y$ rotations and the $Z$-based measurement observables, enhancing the circuit expressivity and enabling exploration of a larger subspace of the Hilbert space.

\subsubsection{Optimization Process}
The variational quantum circuit is optimized to minimize the expectation value of the cost Hamiltonian:
\begin{equation}
\vec{\theta}^* = \arg\min_{\vec{\theta}} \langle\psi(\vec{\theta})|\hat{H}|\psi(\vec{\theta})\rangle
\end{equation}
where $\hat{H}$ is the problem-specific cost Hamiltonian incorporating exploration objectives, environmental constraints, and quantum coherence requirements. The optimization is performed using an adaptive Simultaneous Perturbation Stochastic Approximation (SPSA) algorithm, which efficiently handles the high-dimensional parameter space of the variational circuit while being robust to the statistical noise inherent in quantum measurements.
The adaptive learning rate mechanism adjusts the optimization step size based on progress metrics and stagnation detection:
\begin{equation}
\eta_{t+1} = \eta_t + \mu m_t + (1-\mu)\Delta\eta_t
\end{equation}
where $\mu$ is the momentum coefficient, $m_t$ is the momentum term at iteration $t$, and $\Delta\eta_t$ is the learning rate adjustment based on recent optimization progress. 

\section{Implementation Details}
\label{impl}
\begin{algorithm}
\caption{Hybrid Quantum-Classical NBV System}
\label{alg:hqc-nbv}
\begin{algorithmic}[1]
\REQUIRE $S, v_0, \phi_{FOV}, d_{max}, \tau_{coverage}$
\ENSURE $\mathcal{V} = \{v_0, v_1, \ldots, v_n\}$ : $\mathcal{C}(\mathcal{V}) \geq \tau_{coverage}$
\STATE $\mathcal{V} \gets \{v_0\}$, $\mathcal{M} \gets \text{InitializeMap}(S)$
\STATE $\mathcal{M} \gets \text{UpdateObservation}(\mathcal{M}, v_0)$ 
\WHILE{$\mathcal{C}(\mathcal{M}) < \tau_{coverage}$}
    \STATE $\hat{H} \gets \text{ConstructHamiltonian}(\mathcal{M}, v_t)$ 
    \STATE $U(\vec{\theta}) \gets \text{CreateParameterizedCircuit}(n,L)$ 
    \STATE $|\psi_0\rangle \gets H^{\otimes n}|0\rangle^{\otimes n}$ 
    \STATE $\vec{\theta}_0 \gets \text{InitializeParameters}()$
    \FOR{$i = 0$ \TO $N_{iter}-1$}
        \STATE $c_i \gets \langle\psi(\vec{\theta}_i)|\hat{H}|\psi(\vec{\theta}_i)\rangle$ 
        \STATE $g_i \gets \text{AdaptiveSPSA\_GradientEsti.}(U, \hat{H}, \vec{\theta}_i)$
        \STATE $\eta_i \gets \text{AdaptiveLearningRate}(c_0,...,c_i)$ 
        \STATE $\vec{\theta}_{i+1} \gets \vec{\theta}_i - \eta_i \cdot g_i$ 
    \ENDFOR
    
    \STATE $\vec{\theta}^* \gets \vec{\theta}_{N_{iter}}$
    \STATE $P(x) \gets \{|\langle x|U(\vec{\theta}^*)|\psi_0\rangle|^2\}$ 
    
    \STATE $\vec{q} \gets \text{MajorityVote}(\{x \sim P(x)\})$
    \STATE $v_{next} \gets \text{DecodeParameters}(\vec{q}, v_t, \mathcal{M})$
    
    \STATE $valid \gets \text{ValidateTrajectory}(v_t, v_{next}, \mathcal{M})$
    \IF{$valid$}
        \STATE $v_{t+1} \gets v_{next}$
    \ELSE
        \STATE $v_{t+1} \gets \text{ClassicalFallbackStrategy}(\mathcal{M}, v_t, v_{next})$
    \ENDIF
    
    \STATE $\mathcal{V} \gets \mathcal{V} \cup \{v_{t+1}\}$
    \STATE $\mathcal{M} \gets \text{UpdateObservation}(\mathcal{M}, v_{t+1})$
\ENDWHILE
\RETURN $\mathcal{V}$
\end{algorithmic}
\end{algorithm}

The HQC-NBV system optimizes viewpoint selection for exploring unknown environments with obstacles. The process begins by initializing the scene with an initial viewpoint. In each iteration, before the coverage threshold is reached, the system updates the set of observed points by checking visibility from the current viewpoint. The variational ansatz is initialized for the current viewpoint and the problem Hamiltonian is created that quantifies the cost. The classical cost is calculated, and the parameters are optimized using a Variational Quantum Eigensolver (VQE) with an adaptive SPSA optimizer. The optimal parameters are decoded to determine the next viewpoint, and a trajectory validation is the hard constraint to ensure the new viewpoint within the observed area in $\mathcal{M}$ and does not collide with any obstacles. If the trajectory is not valid, we will select the furthest valid position along the moving direction, named the classical fallback strategy (a detailed description is present in the supplementary material). The process is iteratively executed to find out a sequence of the next viewpoint. The system execution logic can be represented as the algorithm in \ref{alg:hqc-nbv}. 

\section{Experiments and Results}
\label{expAres}
To demonstrate the effectiveness and robustness of the proposed HQC-NBV, we conduct a series of experiments on scenes with different areas and different obstacles. To examine the design of variational ansatz, we also conduct specialized experiments to isolate and quantify the contributions of key quantum components in our hybrid approach aiming to provide insights into how quantum characteristics—specifically entanglement patterns and coherence-preserving terms—impact exploration performance. In this study, the proposed method is implemented using the Qiskit framework, and all the experiments are performed on the Qiskit Aer backend simulator \cite{qiskit}.
The camera parameters used in this study consistently respected a field of view (FOV) $2\pi/3$, and a maximum ray distance of 8 units. The starting view is initialized at a non-collision position.

\subsection{Experimental Setup}
We designed three distinct scenes with varying levels of complexity to comprehensively evaluate the robustness and scalability of the proposed methods. Due to the page limitation, the visualized experimental scenes can be found in the supplementary material.
\begin{itemize}
    \item Scene 1 (S1): Surrounding obstacles in area $20\times20$ $\text{unit}^2$;
    \item Scene 2 (S2): Central obstacle in area $20\times20$ $\text{unit}^2$;
    \item Scene 3 (S3): Complex walls with surrounding obstacles in area $20\times20$ $\text{unit}^2$;
    \item Scene 4 (S4): Surrounding and central obstacles in larger area $40\times40$ $\text{unit}^2$.
\end{itemize}

To investigate the impact of entanglement structure in our approach, we implemented four variants of Ansatz architecture while maintaining identical Hamiltonian formulations and classical optimization procedures: 
\begin{itemize}
    \item Full Architecture (FA): Our proposed bidirectional alternating entanglement pattern with both intra-group and inter-group CNOT gates;
    \item Non-Entangled (NE): A circuit with the same number of parameterized rotations but without any entangling gates, equivalent to independent qubit rotations;
    \item Intra-Group Only (IG): Preserving parameter group coherence through intra-group entanglement but removing connections between different parameter groups;
    \item Inter-Group Only (EG): Maintaining only the connections between parameter groups while removing intra-group entanglement. 
\end{itemize}

To assess the contribution of quantum coherence-preserving terms in our cost Hamiltonian, we conducted a systematic ablation study by modifying the $\hat{H}_{\text{coh}}$ component: 
\begin{itemize}
    \item Complete Hamiltonian (CH): Including all coherence-preserving terms ($X$ and $XX$ operators with adaptive weights);
    \item No Coherence Terms (NC): Removing all $\hat{H}_{\text{coh}}$ components, retaining only the problem-encoding $Z$-based terms;
    \item Single-Qubit X Only (SQX): Preserving the $\sum_i \alpha_{X_i} \hat{X}_i$ terms while removing two-qubit $XX$ interactions.
\end{itemize}

In addition to these, we also comprehensively evaluate the performance of our approach against two classical exploration approaches, RH-NBV and the frontier-based method, regarding the exploration coverage ratio, path length and exploration efficiency.

\subsection{Experimental Results}

\begin{figure}
    \centering
    \includegraphics[width=1.0\linewidth]{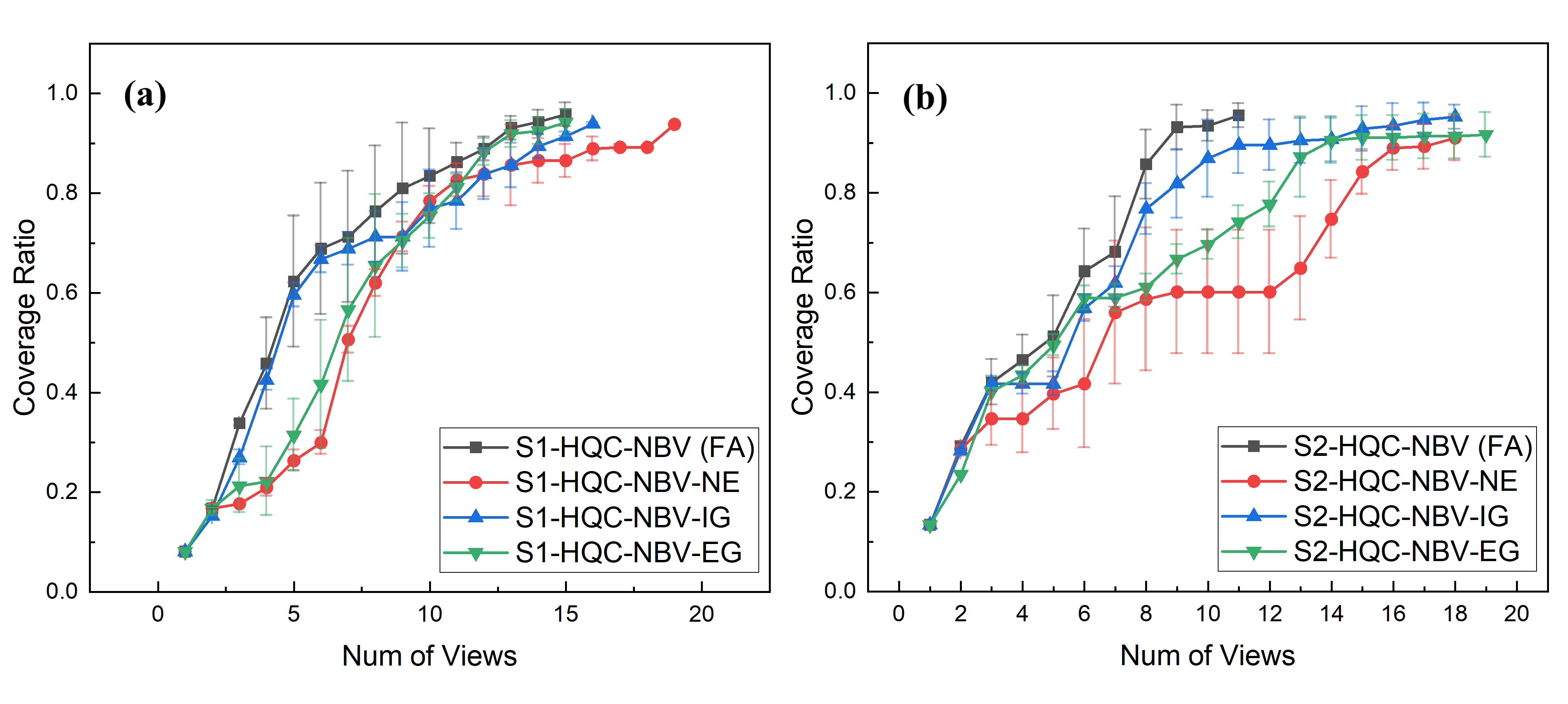}
    \caption{The effectiveness of entanglement architecture on the exploration performance: (a) coverage ratio in Scene 1; (b) coverage ratio in Scene 2.}
    \label{fig:s12_ent}
\end{figure}
Figure \ref{fig:s12_ent} presents the comparative results, demonstrating that the full bidirectional entanglement architecture consistently outperformed reduced-entanglement variants. Notably, the non-entangled circuit required an average of 31.58\% and 52.63\% more views to achieve 90\% coverage in S1 and S2 respectively, highlighting the significant role of quantum correlations in effective exploration planning. The intra-group-only variant performed better than the inter-group-only variant in both scenes, suggesting that maintaining qubits parameter entanglement within logical groups (direction, distance, adjustment, orientation) is more critical than cross-parameter entanglement. The inter-group entanglement architecture also contributes to the improvement compared to the non-entanglement variant because of the intrinsic connection between the logical groups of the informative view planning problem physically. 
\begin{figure}
    \centering
    \includegraphics[width=1.0\linewidth]{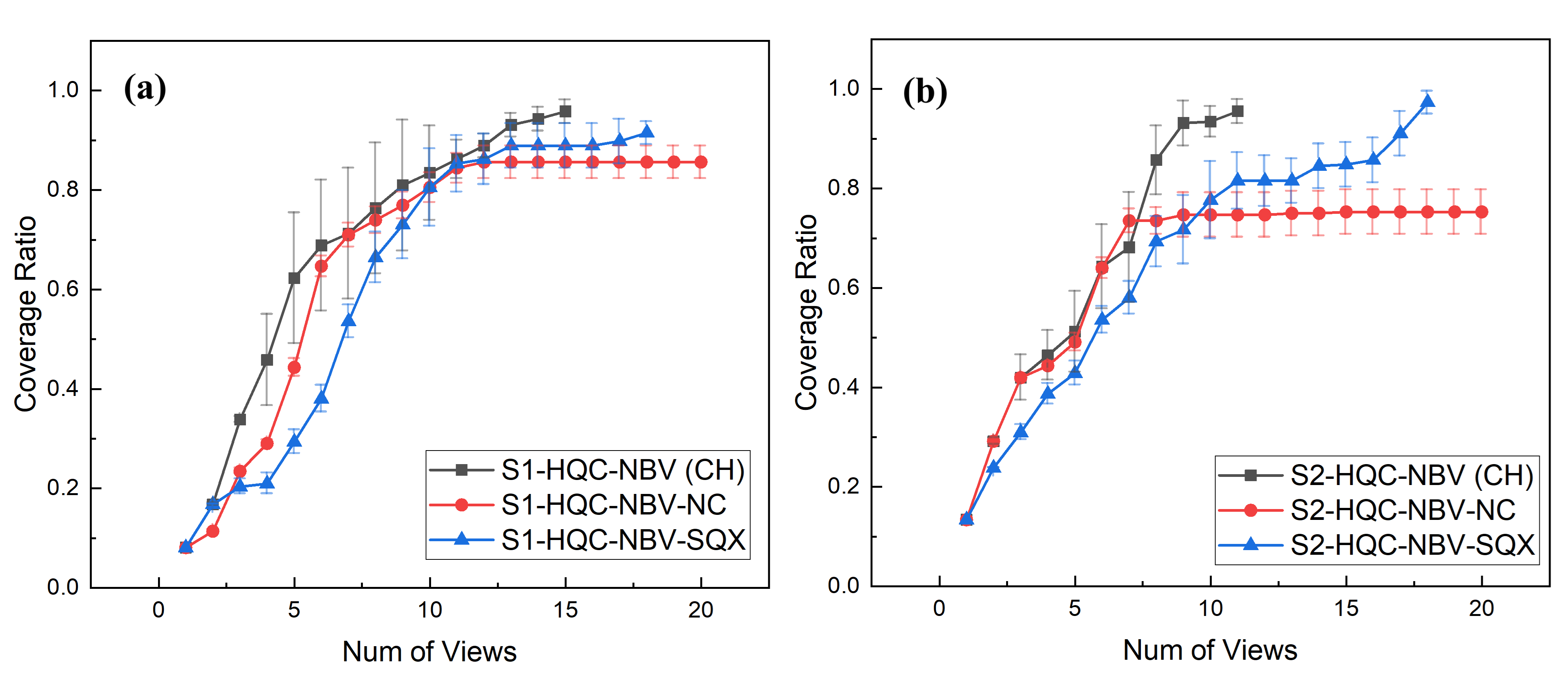}
    \caption{The evaluation of coherence-preserving term on the exploration performance: (a) coverage ratio in Scene 1; (b) coverage ratio in Scene 2.}
    \label{fig:s12_coh}
\end{figure}
\begin{figure}
    \centering
    \includegraphics[width=1.0\linewidth]{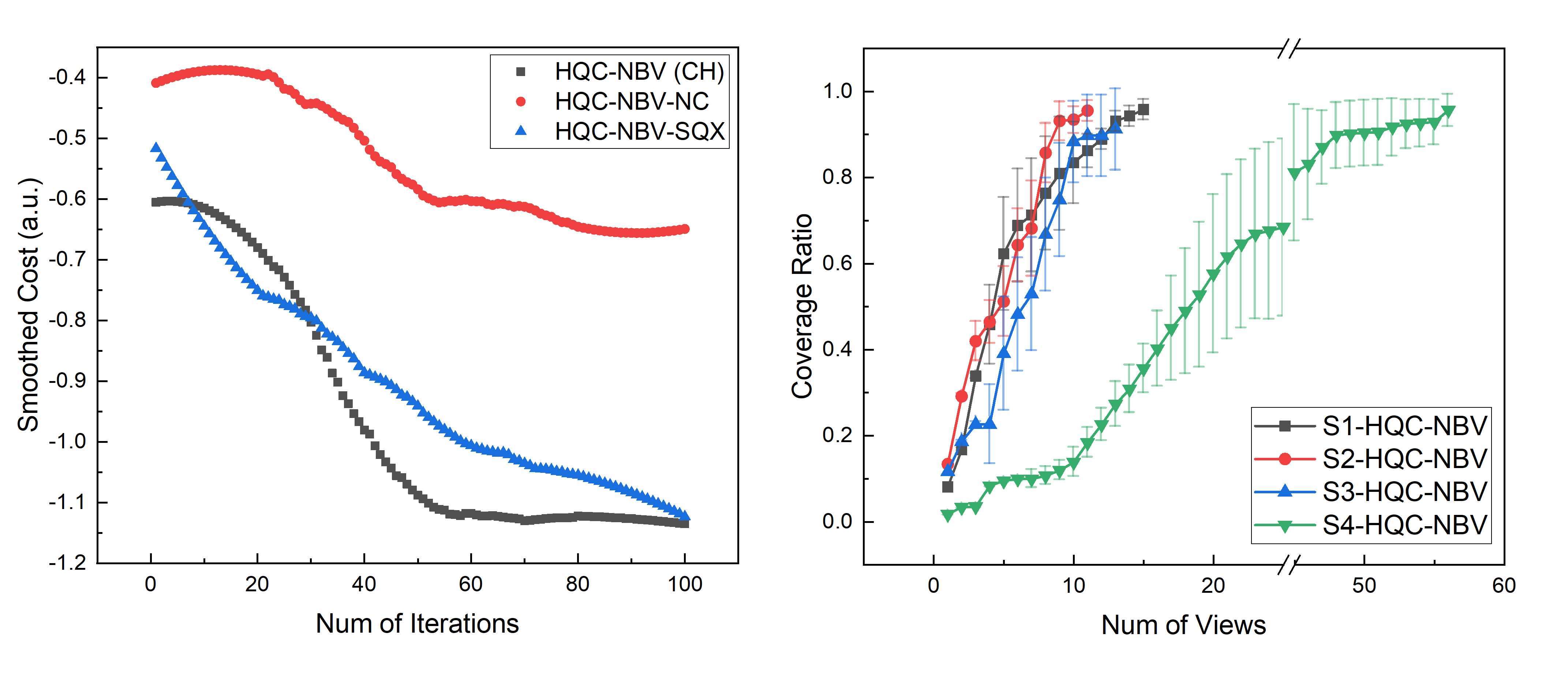}
    \caption{(Left) The smoothed cost against the number of iterations in a single optimization, evaluate the effect of coherence-preserving term on the optimization process; (Right) The coverage against the number of views of our approach in different scenes}
    \label{fig:s1_coh_opt}
\end{figure}
\begin{figure*}
    \centering
    \includegraphics[width=0.95\linewidth]{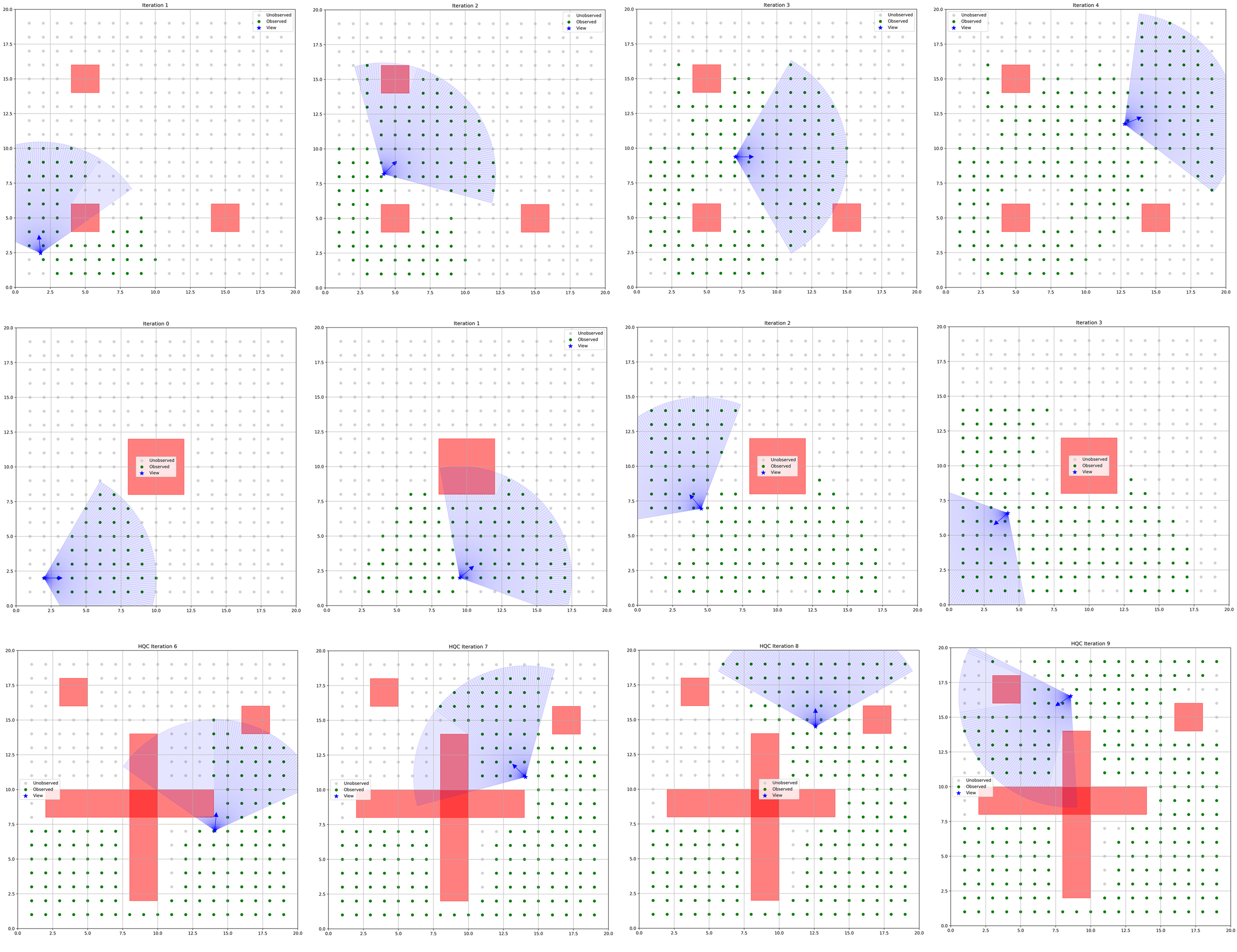}
    \caption{Sample continues views planned by HQC-NBV in S1, S2 and S3. The red rectangles denote the obstacles, the blue wedges represent the FOV of the viewpoint, and the green dots are the observed grid.}
    \label{fig:sample_visual}
\end{figure*}

Figure \ref{fig:s12_coh} illustrates that the absence of coherence-preserving terms led to frequent entrapment in local minima, with the no-coherence variant failing to achieve above 86\% coverage and 75\% in average in Scene 1 and Scene 2, respectively. The performance degradation was most pronounced in later exploration stages (coverage $>$ 70\%), where remaining unexplored regions became sparse and disconnected.
The single-qubit X-only variant demonstrated intermediate performance, maintaining reasonable exploration capabilities but showing reduced ability to escape local minima in complex scenarios. This suggests that while single-qubit superposition maintenance contributes to exploration effectiveness, the two-qubit coherence terms play a crucial role in coordinating parameter updates across different aspects of the navigation decision.

Figure \ref{fig:s1_coh_opt} (Left) presents the convergence behaviour with different coherence-preserving terms configurations during a single representative optimization, showing that the complete Hamiltonian consistently achieved lower final cost values with fewer optimization iterations compared to ablated variants. This demonstrates that the coherence-preserving terms not only improve exploration outcomes but also lead to efficient optimization.

Figure \ref{fig:s1_coh_opt} (Right) demonstrates the robustness and scalability of our approach. Our approach performs an efficient exploration in S1, S2 and S3 within 15 viewpoints. The coverage growth in S2 is more dramatic than that in S1 and S3 due to the simplicity of the scene. The exploration in S4 requires 56 viewpoints to achieve comparable coverage, which is roughly four times the number required for S1, S2 and S3. This scaling factor meets the simple 4:1 ratio of environment sizes (S4 is four times larger in area than S1, S2 and S3), suggesting that the proposed approach does not degrade with the increase in environment size and complexity.

\begin{figure*}
    \centering
    \includegraphics[width=1.0\linewidth]{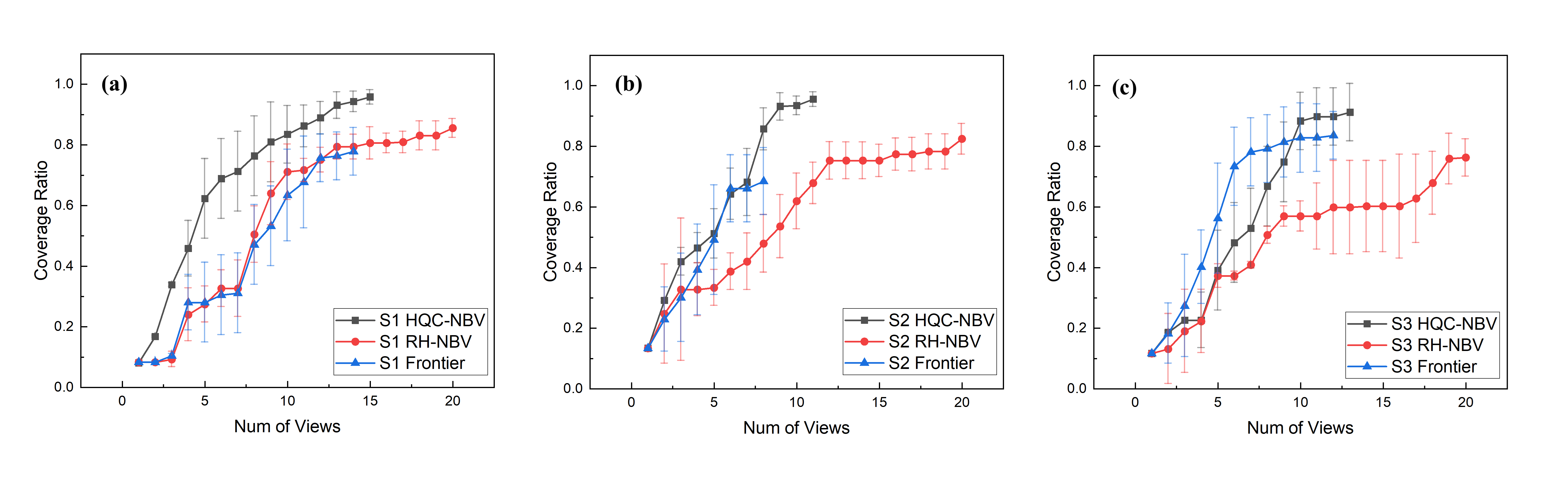}
    \caption{Comparison of coverage ratio progression between HQC-NBV, RH-NBV, and Frontier-based approaches in: (a) Scene 1; (b) Scene 2 and (c) Scene 3.}
    \label{fig:comparison_coverage}
\end{figure*}

\begin{figure}
    \centering
    \includegraphics[width=1.0\linewidth]{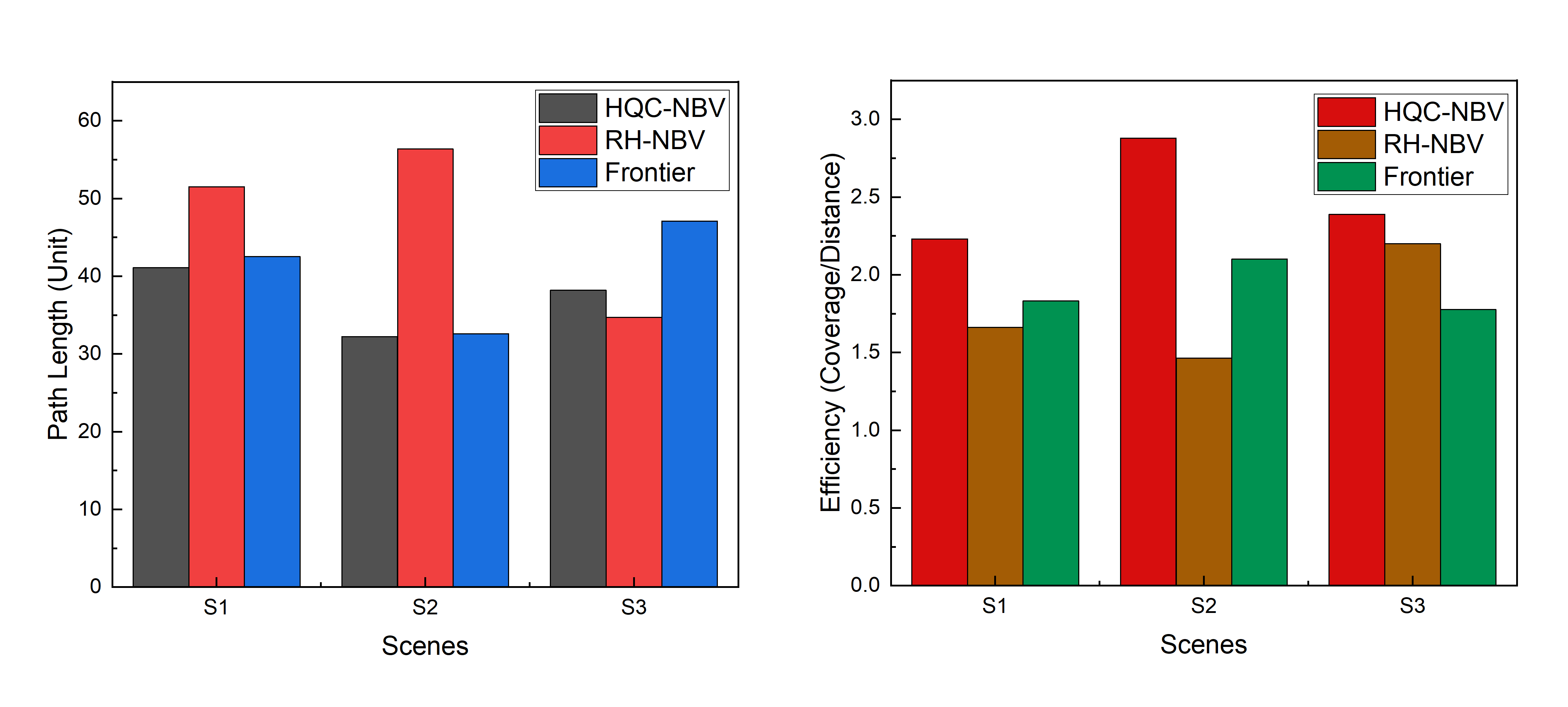}
    \caption{Performance metrics comparison: (a) Total path length across different scenes; (b) Exploration efficiency measured as coverage-to-distance ratio.}
    \label{fig:comparison_metrics}
\end{figure}

To validate the efficacy of our approach against classical methods, we conducted comparative experiments across different scenes. As shown in Figure \ref{fig:comparison_coverage}, HQC-NBV consistently outperformed both RH-NBV and frontier-based approaches in terms of coverage progression. In Scene 1, our method achieved 95.8\% coverage within 16 viewpoints, while classical methods reached a maximum of only 80.6\% coverage with the same number of views. The advantage is even more pronounced in Scene 2, where HQC-NBV reached 95.5\% coverage in just 11 viewpoints, whereas RH-NBV required almost twice as many viewpoints to achieve 82.4\% coverage. In the most challenging environment Scene 3, the frontier-based approach demonstrated superior performance during the initial exploration phase, this early advantage can be attributed to the method's greedy frontier selection strategy, which excels at covering open spaces quickly. However, as shown in Figure \ref{fig:comparison_coverage}(c), the frontier-based approach encountered early termination around 83.6\% coverage due to the method's inherent limitations in navigating complex environments with disconnected unexplored regions. Our approach maintained consistent progress throughout the exploration process, ultimately achieving 92.2\% coverage within 13 viewpoints—significantly outperforming both classical approaches. These experiments highlight the robustness of our method in handling exploration scenarios with different complexities, particularly where classical approaches face significant limitations.

The quantitative comparison in Figure \ref{fig:comparison_metrics} further demonstrates the efficiency of our approach. It consistently generated shorter exploration paths, with 20.2-42.8\% reduction in total path length compared to RH-NBV across all scenes. This efficiency is particularly evident in Scene 2, where HQC-NBV's path length (32 units) was significantly lower than RH-NBV (56 units). The efficiency metric (coverage-to-distance ratio) confirms our method's superiority, showing 7.9-49.2\% higher exploration efficiency than classical approaches.

Notably, Figure \ref{fig:comparison_coverage} also indicates that HQC-NBV exhibits more stable performance across multiple runs, particularly during the critical middle phase of exploration (viewpoints 5-10). This stability highlights the robustness of our quantum-enhanced approach in handling exploration uncertainty, especially in complex environments like Scene 3, where classical methods showed considerable variability in performance.

Figure \ref{fig:sample_visual} presents two groups of continuous views planned by HQC-NBV in S1, S2 and S3, respectively. It demonstrates the effectiveness of our viewpoint planning algorithm in progressively expanding coverage across different scenes, over several iterations. The samples start from an initial status where only a few areas have been explored, the algorithm efficiently selects viewpoints that maximize the coverage of unobserved regions. With each subsequent iteration, the coverage area grows substantially with feasible movement. Due to the page limitation, more visualized experimental results are presented in the supplementary materials.


\section{Conclusion}
\label{conc}
In this paper, we present a paradigm-shifting scheme in view planning, namely, Hybrid Quantum-Classical Next-Best-View (HQC-NBV) for autonomous exploration tasks. Our approach features a multi-component quantum Hamiltonian and a variational circuit with bidirectional entanglement patterns. Experiments across various environments demonstrated that quantum-specific elements provide measurable contributions, with our entanglement architecture and coherence-preserving terms significantly enhancing exploration efficiency. Compared to the classical approaches, our method consistently achieved higher coverage rates (up to 95.8\%) with 7.9-49.2\% higher exploration efficiency against travel lengths. Moreover, our approach demonstrated excellent scalability and robustness across environments of increasing size and complexity. The framework achieves high-efficiency exploration while being compatible with current NISQ devices. This work paves the first step toward integrating quantum variational algorithms for solving robot vision problems. 

{
    \small
    \bibliographystyle{ieeenat_fullname}
    \bibliography{main}
}

\end{document}